\documentclass[journal]{IEEEtran}

\usepackage{multirow}
\usepackage[table,xcdraw]{xcolor}
\usepackage{graphicx}
\usepackage{amsmath}
\usepackage{amsfonts}
\usepackage{algorithm}
\usepackage{amsthm,amssymb}
\usepackage{mathrsfs}
\usepackage{algorithmic}
\usepackage{makecell}
\usepackage{easyReview}
\usepackage{booktabs}
\setcellgapes{3pt}

\setcellgapes{3pt}

\usepackage[justification=centering]{caption}
\usepackage{color}
\usepackage[subfigure]{tocloft}
\usepackage{subfigure}
\usepackage{tabularx}
\usepackage{amsmath, bm}
\usepackage{hyperref}
\hypersetup{hidelinks,
	colorlinks=true,
	allcolors=blue,
	pdfstartview=Fit,
	breaklinks=true}

\usepackage{cite}

\ifCLASSINFOpdf
\else
\fi
\begin{document}

\title{Explainable Semantic Federated Learning Enabled Industrial Edge Network for Fire Surveillance}
\author{Li Dong, Yubo Peng, Feibo Jiang, \textit{Senior Member, IEEE},  Kezhi Wang, \textit{Senior Member, IEEE}, and Kun Yang, \textit{Fellow, IEEE}
\thanks{
		This work was supported in part by the
		National Natural Science Foundation of China under Grant 41904127 and 62132004, in part by the Hunan Provincial Natural Science Foundation of China under Grant 2024JJ5270, in part by the Open Project of Xiangjiang Laboratory under Grant 22XJ03011, and in part by the Scientific Research Fund of Hunan Provincial Education Department under Grant 22B0663.
		(Corresponding authors: Yubo Peng, Feibo Jiang.)
		
		Li Dong (Dlj2017@hunnu.edu.cn) is with the School of Computer Science, Hunan University of Technology and Business, Changsha 410205, China, and also with the Xiangjiang Laboratory, Changsha 410205, China.
		
		Yubo Peng (pengyubo@hunnu.edu.cn) is with the School of Intelligent Software and Engineering, Nanjing University, Suzhou, China.
		
		Feibo Jiang (jiangfb@hunnu.edu.cn) is with the Hunan Provincial Key Laboratory of Intelligent Computing and Language Information Processing, Hunan Normal University, Changsha 410081, China.

		Kezhi Wang (Kezhi.Wang@brunel.ac.uk) is with the Department of Computer Science, Brunel University London, UK.
			
		Kun Yang (kyang@ieee.org) is with the School of Intelligent Software and Engineering, Nanjing University, Suzhou, China.
}
}
\markboth{Submitted for Review}%
{Shell \MakeLowercase{\textit{et al.}}: Bare Demo of IEEEtran.cls for IEEE Journals}
%



\maketitle

\begin{abstract}
	In fire surveillance, Industrial Internet of Things (IIoT) devices require transmitting large monitoring data frequently, which leads to huge consumption of spectrum resources.
	Hence, we propose an Industrial Edge Semantic Network (IESN) to allow IIoT devices to send warnings through Semantic communication (SC). Thus, we should consider (1) Data privacy and security. (2) SC model adaptation for heterogeneous devices. (3) Explainability of semantics. 
	Therefore, first, we present an eXplainable Semantic Federated Learning (XSFL) to train the SC model, thus ensuring data privacy and security.
	Then, we present an Adaptive Client Training (ACT) strategy to provide a specific SC model for each device according to its Fisher information matrix, thus overcoming the heterogeneity. 
	Next, an Explainable SC (ESC) mechanism is designed, which introduces a leakyReLU-based activation mapping to explain the relationship between the extracted semantics and monitoring data. Finally, simulation results demonstrate the effectiveness of XSFL.
\end{abstract}


\section{Introduction}
Every year, industrial fires pose a serious threat to human lives and property due to human errors or system failures. The risk of fire is further heightened by high infrastructure concentration and stored inventory \cite{prabha2019iot}. When a fire occurs, it is crucial to promptly issue warnings and initiate rescue missions to minimize property damage and increase the chances of saving lives. However, traditional manual monitoring methods are inconvenient, scattered, and come with certain drawbacks such as high costs and poor network connectivity \cite{thekkil2017real}. Therefore, effectively and timely sending critical information upon detecting a fire presents a significant challenge for traditional manual monitoring methods.

With the development of Artificial Intelligence (AI) and the Industrial Internet of Things (IIoT), AI-empowered IIoT devices play an important role in catering to a multitude of problems (e.g., transportation, healthcare, and disaster management, etc.) efficiently and reliably \cite{jiang2020ai}. Simultaneously, it is also used for fire surveillance. 
In \cite{muhammad2019efficient}, the authors proposed an efficient system, which was based on a lightweight Deep Learning (DL) model, for fire detection in uncertain surveillance scenarios.
In \cite{9531399}, the authors introduced a vision-based fire detection framework that focuses on monitoring private spaces while prioritizing occupant privacy.
An innovative solution for continuous forest surveillance and fire detection was developed in \cite{manikandan2022design} using a drone-based system integrated with AI capabilities, specifically designed for forest fire-fighting purposes.
In \cite{10084949}, a novel approach leveraging IIoT technology was proposed to detect fires and promptly notify relevant parties. This method ensured accurate solutions for fire surveillance systems.


However, IIoT devices require frequent transmissions of large monitoring data when the conflagration occurs, leading to a huge consumption of spectrum resources.
Semantic Communication (SC) is viewed as one of the intelligent solutions to physical layer communication.
Different from traditional ways, SC transmits only semantic features required for the respective task \cite{10558819}. The semantic feature is generally defined as the symbols with much fewer dimensions extracted from the raw data.
For example, in fire surveillance, the IIoT devices only extract the semantic information from the surveillance image with the deployed SC encoder, then transmit it to the Edge Server (ES) deployed on the Base Station (BS). The ES decodes the received semantic information with the deployed SC decoder and obtains the prediction results.
Since the transferred data is greatly reduced in SC, the consumption of spectrum resources and communication delay are also greatly decreased. The high-accuracy SC model based on DL is critical for the performance of SC, while several challenges should be considered as follows:
\begin{enumerate}[]
	\item {\it{Data privacy and security for industrial applications:}}
	Most data in IIoT devices are sensitive for referring personal information and business secrets \cite{yang2022privacy}. Hence, privacy security has become a critical issue.
	However, all data is required to be stored centrally for fire surveillance, and the detection model is trained and carried out in the traditional centralization method, which brings overwhelming communication costs and huge risks of data leakage for edge users.
	
	\item {\it{High training delay for heterogeneous devices:}} In the edge network, the IIoT devices are heterogeneous. For well-resourced devices, a sophisticated global model could ensure the model's accuracy. However, such a global model may bring a high training delay for the limited-resourced devices, which is not accepted for fire surveillance \cite{chen2013natural}. Although a compact global model could ensure the training delay for all participating IIoT devices, the model accuracy is difficult to guarantee, especially for well-resourced IIoT devices \cite{yang2019federated}.  
	
	\item {\it{Explainability of extracted fire semantics for succors:}}
	The erroneous predictions by DL models may potentially result in fatal accidents in recent years. Hence, succors couldn't trust DL when making life-critical decisions for it is nontransparent and unexplainable \cite{begoli2019need}. Therefore, the DL-based extracted semantics, representing the features of fire, must be transparent and explainable.   
\end{enumerate}

To address the above issues, we propose an Industrial Edge Semantic Network (IESN) that reduces the transmitted information and thus decreases the communication overheads of IIoT devices. Our main contributions can be summarized as follows:

\begin{enumerate}[]
	\item We introduce the architecture of IESN, in which the SC is used as the communication way between IIoT devices and ES. SC can make devices transmit only the extracted semantic information rather than the raw data. Thus, the IESN can reduce the transmitted information and improve the communication efficiency in edge networks. Furthermore, we propose an eXplainable Semantic Federated Learning (XSFL) for IIoT devices and ES to co-train the SC model, and this decentralized learning method at the edge can protect the privacy and security of industrial data.

	\item We present the Adaptive Clients Training (ACT) strategy in the ES. Specifically, we first cluster the IIoT devices according to their local resources. Then, for each IIoT device, ES publishes a specific FL model with a proportion of frozen parameters according to the Fisher information of its local weights. The fewer local resources the IIoT device owns, the more the frozen ratio of the parameters, and thus the FL training delay is decreased. Therefore, the local FL models can better adapt to heterogeneous devices with different resources, and the second challenge is solved.

	\item We propose the Explainable SC (ESC) mechanism in edge monitor devices to provide explainability of the extracted semantics based on DL. Concretely, the ESC uses semantic-specific gradient information to localize important semantic features. Then, we introduce a leakyReLU-based activation mapping to combine the localized features with the raw monitoring data to develop a high-resolution heatmap visualization. Thus, the decision-makers can understand the relative between the monitoring data and extracted semantics. Therefore, we address the third challenge.
\end{enumerate}

The rest of this paper is organized as follows. The system model is introduced in Section II. The proposed XSFL is described in Section III. The complexity analyses are introduced in Section IV. Section V presents the numerical results. The work summary and future works are described in Section VI.

\section{System Model}
\begin{figure}[htbp]
	\centering
	\subfigure[]{
		\includegraphics[width=8.5cm]{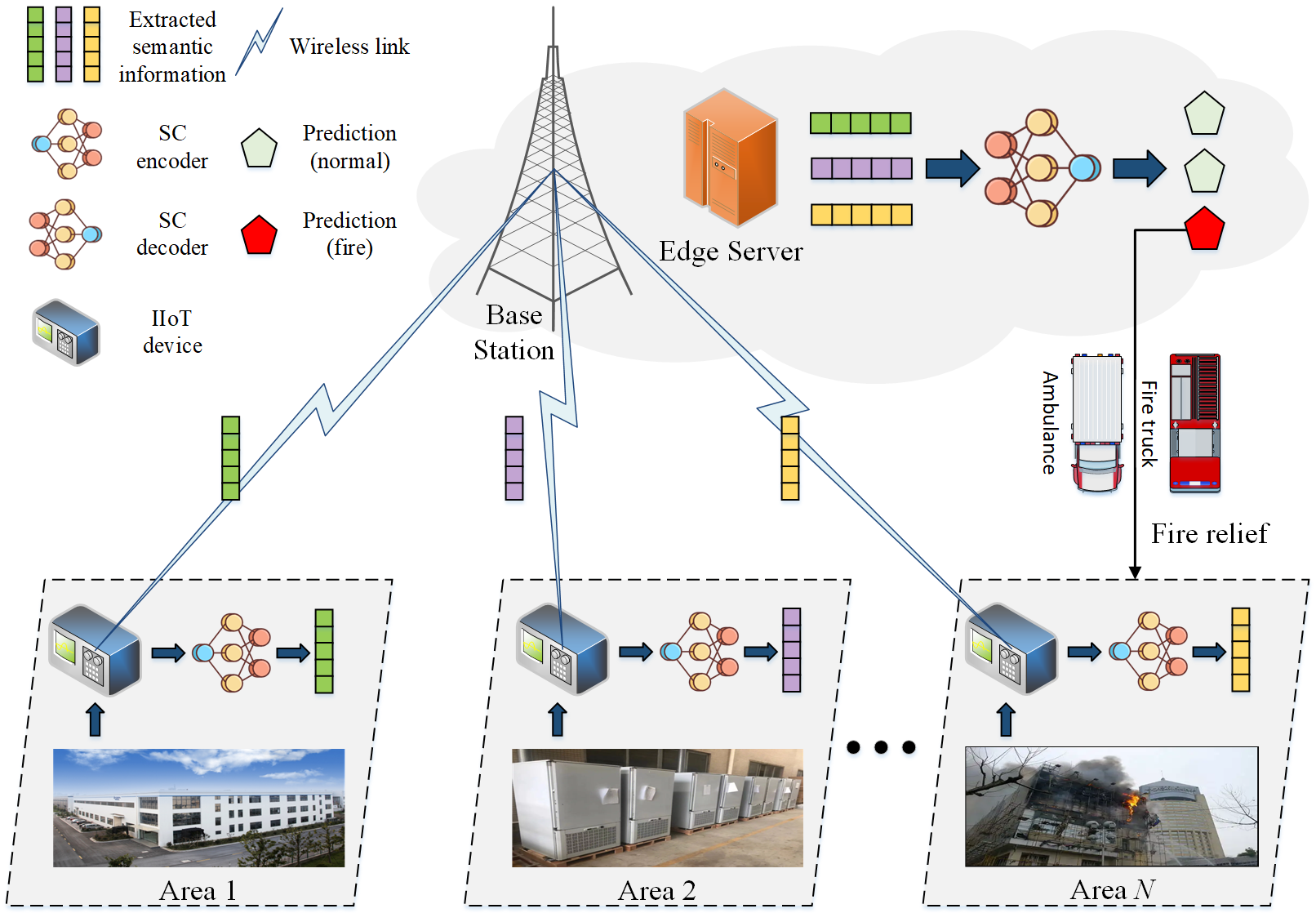}
	}
	\quad
	\subfigure[]{
		\includegraphics[width=8.5cm]{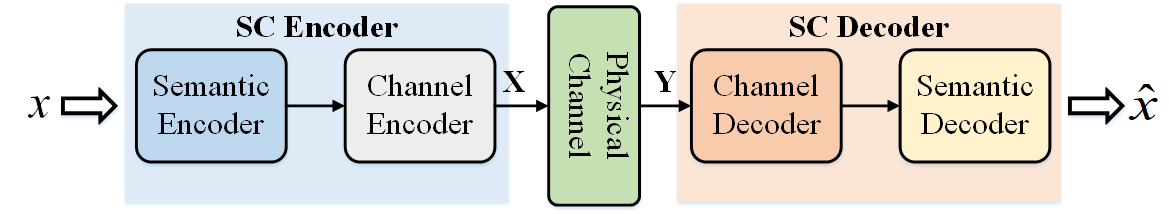}
	}
	\quad
	\caption{The illustration of IESN for fire surveillance. (a) Distributed SC system over the wireless network. (b) The architecture of the SC model.}
	\label{fig:system}
\end{figure}

Fig. \ref{fig:system} (a) shows the architecture of the proposed IESN in fire surveillance. For a spectrum resource-limited wireless network, we deploy a distributed SC system which consists of $N$ IIoT devices, and one ES deployed on the BS. The set of IIoT devices is denoted as $\mathcal{N}=\{1,2,...,N\}$.
The IIoT devices only need to transmit the semantic information to the BS, rather than the raw monitoring data, then the semantic information is decoded in the ES. 
Once the decoding results of the received semantics represent the occurrence of a fire, the decision-makers should take the corresponding actions, e.g., dispatching ambulances and fire trucks. To enable IIoT devices can extract the semantic information, the SC encoder is deployed on each IIoT device, while the SC decoder is deployed on the ES to decode the received semantic information. The architecture of the SC model is introduced in Section III-A. In addition, we should consider the impairments of the physical channel between IIoT devices and the BS.
\subsection{Semantic Communication Model in IESN}
The encoding and decoding of the semantic information are critical for the accuracy of SC, hence we must consider the architecture design of the SC model for IESN.
As shown in Fig. \ref{fig:system} (b), the SC model mainly consists of three parts: SC encoder, physical channel, and SC decoder. The SC encoder includes the semantic and channel encoder, and it is deployed on each IIoT device. The SC decoder consists of the channel and semantic decoder, and it is deployed on ES.
Note the above parts of the SC model are built based on deep neural networks, and different network layers (e.g. convolutional layer, transformer layer, etc.) can be used according to different semantic tasks \cite{wang2022transformer}.

The SC encoder encodes the input raw data $x$ and extracts the semantic information. The encoding result can be expressed as:
\begin{equation}\label{eq:Shi1}
	\mathbf{X} = C\left(S\left(x,\bm{\vartheta}\right),\bm{\alpha}\right),
\end{equation}
where $\mathbf{X}$ represents the extracted semantic information; $S\left(\cdot\right)$ represents the semantic encoder with model parameters $\bm{\vartheta}$ and $C\left(\cdot\right)$ is the channel encoder with model parameters $\bm{\alpha}$. 

When transmitted over the physical channel, $\mathbf{X}$ suffers transmission impairments that include distortion and noise. The transmission process can be given by:
\begin{equation}\label{eq:Shi2}
	\mathbf{Y} = \mathbf{hX+N},
\end{equation}
where $\mathbf{Y}$ represents the received semantic information; $\mathbf{h}$ represents the channel gain between the transmitter and the receiver; $\mathbf{N}$ is Additive White Gaussian Noise (AWGN). For end-to-end training of the encoder and decoder, the physical channel must allow backpropagation. Therefore, the physical channel can be simulated by neural networks \cite{park2020end,10398266}. 

The SC decoder performs the decoding of the received semantic information $\mathbf{Y}$ and alleviates transmission impairments. The decoding result $\hat{x}$ can be given by:
\begin{equation}\label{eq:Shi3}
	\hat{x}=S^{-1}\left(C^{-1}(\mathbf{Y}, \bm{\beta}), \bm{\delta}\right),
\end{equation}
where $C^{-1}\left(\cdot\right)$ represents the channel decoder with model parameters $\bm{\beta}$; $S^{-1}\left(\cdot\right)$ is the semantic decoder with model parameters $\bm{\delta}$.

In this paper, we approach the SC task as image classification. As such, we utilize the cross-entropy (CE) loss function in the SC system:
\begin{equation}\label{eq:Shi4}
	\mathcal{L}_{\mathrm{CE}}(\mathbf{y}, \mathbf{\hat{y}})=-\sum_{i=1}^{M} y_{i} \log \left(\hat{y}_{i}\right),
\end{equation}
where $\mathbf{y}=[y_1,y_2,..., y_M]$ represents the array of labels, if the input data $x$ belongs to the $i$-th class, then $y_i=1$ otherwise 0; $\mathbf{\hat{y}}=[\hat{y}_1,\hat{y}_2,..., \hat{y}_M]$ represents the array of the probabilities of all classes, $\hat{y}_i$ represents the probability that $x$ is predicted as the $i$-th class; $M$ is the total number of categories. 
Moreover, classification-related metrics, such as accuracy, precision, etc., can be used to evaluate the quality of SC \cite{gunduz2022beyond}.

\subsection{Federated Learning Model in IESN}
We denote the local dataset of the IIoT device $n$ ($n \in \mathcal{N}$) as $\mathcal{D}_n=\left\{ (x_{n,1},y_{n,1}), (x_{n,2},y_{n,2}), ..., (x_{n,D_n},y_{n,D_n})\right\}$, where $D_n$ is the number of collected samples in $\mathcal{D}_n$, $x_{n,i}$ is the $i$-th sample and $y_{n,i}$ is the corresponding label. 

In the $t$-th communication round of FL, the local loss function of the IIoT device $n$ can be calculated as:
\begin{align}\label{eq:Shi5}
	{F_{n}}\left( \mathbf{w}_{n,t} \right) = \frac{1}{D_n}{\sum\limits_{i=1}^{D_n}{f\left( \mathbf{w}_{n,t}, x_{n,i},y_{n,i}\right)}},
\end{align}
where $f(\mathbf{w}_{n,t},x_{n,i},y_{n,i})$ is the loss function for the $i$-th sample $\left(x_{n,i},y_{n,i} \right)$ in $\mathcal{D}_{n}$; $\mathbf{w}_{n,t}$ represents the weight assigned to the local FL model of the IIoT device $n$ in the $t$-th communication round. Moreover, $\mathbf{w}_{n,t}$ includes all the SC parameters, namely $\mathbf{w}_{n,t}=(\bm{\alpha}_{n,t},\bm{\vartheta}_{n,t},\bm{\beta}_{n,t}, \bm{\delta}_{n,t})$, where $\bm{\alpha}_{n,t},\bm{\vartheta}_{n,t},\bm{\beta}_{n,t},\bm{\delta}_{n,t}$ represent the parameters of the SC model of the IIoT device $n$ from the channel encoder to the semantic decoder in the $t$-th communication round.

To ensure data privacy and security, the global FL model is updated in the ES by aggregating the weights $\mathbf{w}_{n,t}$ of the local FL models from various IIoT devices. The update of the global FL model can be calculated by combining and processing these weights as follows:
\begin{align}\label{eq:Shi6}
	\mathbf{w}_{g,t}=\frac{1}{D}\sum\limits_{n=1}^{N} D_{n} \mathbf{w}_{n,t},
\end{align}
where $\mathbf{w}_{g,t}$ is the updated weight of the global FL model; $D=\sum\limits_{n=1}^{N} D_{n}$ represents the sum of local data. Note that $\mathbf{w}_{g,t}$ shares the same architecture as $\mathbf{w}_{n,t}$.

FL aims to minimize the local FL loss of all devices with an optimal global model $\mathbf{w}_{g,t}$, thus achieving global optimization. Hence, the global loss function of FL can be given by:
\begin{equation}\label{eq:Shi7}
	F_{g}\left( \mathbf{w}_{g,t} \right) = \frac{1}{N}{\sum\limits_{n = 1}^{N}{F_{n}\left(\mathbf{w}_{g,t}\right)}},
\end{equation}
where $F_{n}\left(\mathbf{w}_{g,t}\right)$ represents that the local FL loss based on $\mathbf{w}_{g,t}$. In this paper, the minimization of $F_{g}\left(\mathbf{w}_{g,t}\right)$ is the training goal of the SC model in IESN.

\subsection{Transmission Model in IESN}
In IESN, the uplink data rate of the IIoT device $n$  via the channel $j$ can be calculated as \cite{wei2021low}:
\begin{equation}\label{eq:tp1}
	R_{n, j}^{\mathrm{U}}(t)=B^{\mathrm{U}} \log _{2}\left\{1+\frac{P^{\mathrm{U}} h_{n, j}(t)}{I_{n, j}^{\mathrm{U}}(t)+\sigma}\right\},
\end{equation}
where $P^{\mathrm{U}}$ denotes the transmission power of the IIoT device; $h_{n, j}(t)$ represents the channel gain when device $n$ communicates with the ES via channel $j$; $I_{n, j}^{\mathrm{U}}(t)$ refers to the co-channel interference experienced; $\sigma$ indicates the noise power spectral density; and $B^{\mathrm{U}}$ signifies the bandwidth utilized during the uploading process.

Similarly, when the global model is transmitted to the device $n$ by the ES, the data rate of the downlink can be expressed as:
\begin{equation}\label{eq:tp2}
	R_{n, j}^{\mathrm{D}}(t)=B^{\mathrm{D}} \log _{2}\left\{1+\frac{P^{\mathrm{D}} h_{n, j}(t)}{I_{n, j}^{\mathrm{D}}(t)+\sigma}\right\},
\end{equation}
where $P^{\mathrm{D}}$ represents the transmission power of the ES; $B^{\mathrm{D}}$ signifies the bandwidth of the downlink channel; and $I_{n, j}^{\mathrm{D}}(t)$ denotes the interference experienced during the downloading process.

Hence, the uplink and downlink transmission delays can be respectively derived:
\begin{equation}\label{eq:tp3}
	d_{n,j}^{\mathrm{U}}(t)=\frac{Z\left(\mathbf{w}_{n,t}\right)}{R_{n, j}^{\mathrm{U}}(t)},~~~
	d_{n,j}^{\mathrm{D}}(t)=\frac{Z\left(\mathbf{w}_{n,t}\right)}{R_{n, j}^{\mathrm{D}}(t)},
\end{equation}
where  $Z\left(\mathbf{w}_{n,t}\right)$ denotes required bits for transmitting $\mathbf{w}_{n,t}$ over wireless links.

\subsection{Computation Model in IESN}
We presume that each IIoT device is furnished with a CPU to perform local training tasks. The computational capacity of each device is assessed by its CPU frequency, denoted as $f_{n,t}$. In addition, $\Phi_n(\mathbf{w}_{n,t})$ represents the number of CPU cycles required to carry out the forward-backward propagation process utilizing a single data point for the local model $\mathbf{w}_{n,t}$. Consequently, for the $t$-th communication round, the calculation delay of the local FL model can be derived \cite{wei2021low}:
\begin{equation}\label{eq:cp1}
	d_{n}^{\mathrm{L}}(t)=\frac{D_{n} \Phi_n(\mathbf{w}_{n,t})G}{f_{n,t}},
\end{equation}
where $G$ is the local training iterations.

\subsection{Delay Analysis and Optimization Objective}
Assuming the ES possesses ample computational resources, the delay caused by global model aggregation is considered insignificant \cite{wei2021low}. Therefore, the FL training delay for the $n$-th device during the $t$-th communication round can be expressed as follows:
\begin{equation}\label{eq:pf1}
	d_{n,t} = d_{n,j}^{\mathrm{U}}(t)+d_{n,j}^{\mathrm{D}}(t)+d_{n}^{\mathrm{L}}(t).
\end{equation}

Since the global aggregation step is synchronous in FL \cite{10091843}, during the $t$-th communication round, the total delay of the FL training can be expressed as:
\begin{equation}\label{eq:pf2}
	d_t=\max _{n \in \mathcal{N}}\left\{\min \left\{d_{n, t}, d_{\max}\right\}\right\},
\end{equation}
$d_{\max}$ signifies the maximum permitted delay for each IIoT device $n$ when executing the FL algorithm. Specifically, if $d_{n, t} > d_{\max}$, IIoT device $n$ cannot participate in the $t$-th round of FL training. The primary objective of this paper is to minimize both the global FL loss and the FL training delay.

\section{The XSFL Framework for IESN}
To achieve the IESN for fire surveillance, a high-accuracy SC model is critical. 
Hence, the XSFL framework is proposed, in which we apply the ACT strategy and ESC mechanism in the ES and IIoT devices, respectively.

\subsection{XSFL Framework}
\begin{figure*}[htbp]
	\centering
	\includegraphics[width=18cm]{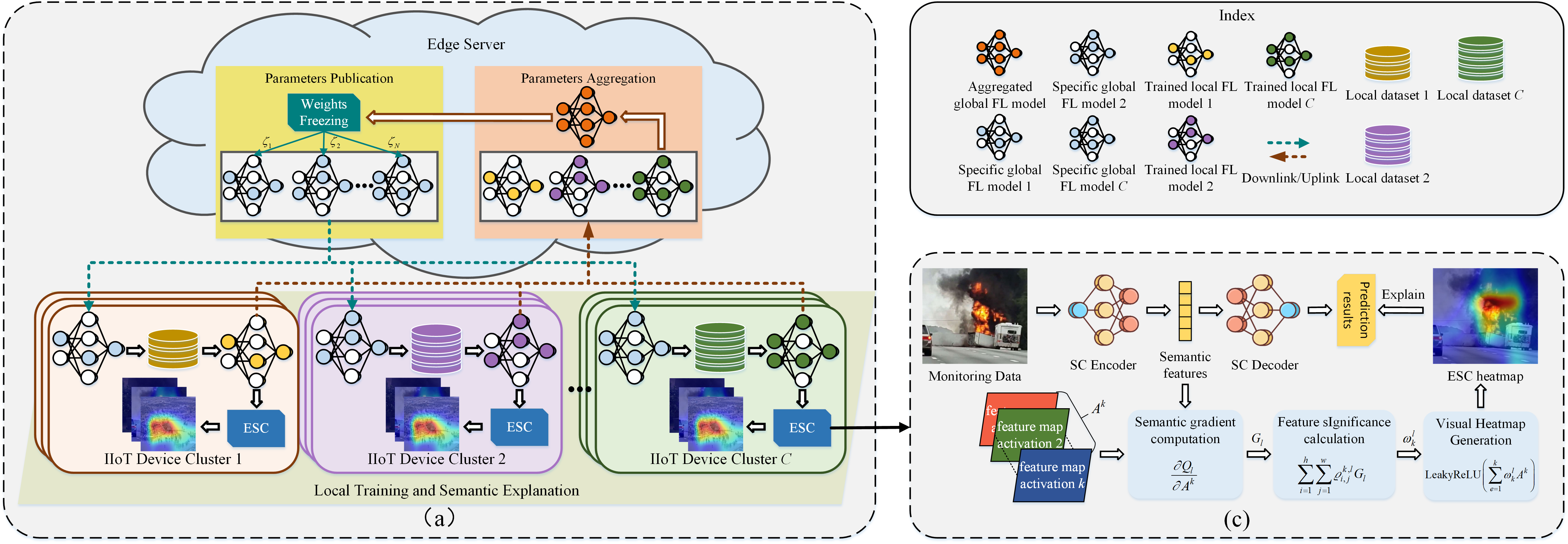}
	\caption{The illustration of the XSFL framework. (a) ACT strategy; (b) ESC mechanism.}
	\label{fig:XSFL}
\end{figure*}
In this subsection, we introduce the details of the XSFL framework. 
We assume that XSFL starts from a certain round $t$, as shown in Fig. \ref{fig:XSFL}, the workflow of the XSFL is as follows:
\subsubsection{Parameters Publication}
To make the local FL models adaptive to the heterogeneous IIoT devices, we perform the ACT strategy on the global FL model $\mathbf{w}_{g,t}$ in ES and the global parameters that require training for each IIoT device are published. Afterward, according to the published parameters, the global FL models are processed specifically, and they are broadcasted to the corresponding IIoT devices for updating and starting the $t+1$-th FL:
\begin{equation}\label{eq:flup}
	\mathbf{w}_{n,t+1}=\hat{\mathbf{w}}_{g,t+1}^n,
\end{equation}
where $\mathbf{w}_{n,t+1}$ denotes the local FL model during the $(t+1)$-th communication round, while $\hat{\mathbf{w}}_{g,t+1}^n$ represents the global FL model specific to IIoT device $n$. It is important to note that both $\mathbf{\hat{w}}_{g,t+1}^n$ and $\mathbf{w}_{n,t+1}$ share the same architectural design.

\subsubsection{Local Training and Semantic Explanation}
Each IIoT device $n$ performs local training with the local FL model $\mathbf{w}_{n,t+1}$.
After training, each IIoT device applies the ESC mechanism to explain the fire semantics from the local FL model in a visual way, which reveals the mapping between semantic features extracted by the SC encoder and input monitoring data. 

\subsubsection{Parameters Aggregation}
When the trained local FL models from all IIoT devices are uploaded to the ES, the ES performs parameters aggregating. Afterward, we obtain the aggregated global FL model. Assuming the total number of communication rounds is $T$, the local training epochs of each IIoT device is $G$, and the proposed XSFL framework is summarized in \textbf{Algorithm} \ref{alg:XSFL}.

\begin{algorithm}
	\caption{XSFL framework}
	\label{alg:XSFL}
	\begin{algorithmic}[1]
		\REQUIRE $T$, $G$.
		\ENSURE $\mathbf{w}_{g,T}$.
		\FOR{each communication round $t \in T$}
		\STATE{\textbf{Edge server do}}
		\STATE{Aggregate local FL models from IIoT devices according to Eq. (\ref{eq:Shi6}).}
		\STATE{Perform ACT strategy on the global FL model according to \textbf{Algorithm} \ref{alg:ACT}.}
		\STATE{Broadcast the specific global FL models $\hat{\mathbf{w}}_{g,t+1}^n$ corresponding to each IIoT device for updating according to Eq. (\ref{eq:flup}).}
		\STATE{\textbf{Edge device do}}
		\FOR{each local epoch in $G$}
		\STATE{Update $\mathbf{w}_{n,t}$ by minimizing Eq. (\ref{eq:Shi5}).}
		\ENDFOR
		\STATE{Explain the extracted semantics with the ESC mechanism according to \textbf{Algorithm} \ref{alg:ESC}.}
		\ENDFOR
	\end{algorithmic}
\end{algorithm}


\subsection{Adaptive Clients Training Strategy}
Practically, IIoT devices are usually heterogeneous, which makes different IIoT devices with disparate local resources have different FL training delays. From Eq. (\ref{eq:tp3}) and Eq. (\ref{eq:cp1}), it becomes evident that the FL training delay is affected by the size of the FL model parameters. Moreover, the model size also plays a significant role in influencing the model's accuracy. Hence, there exists a tradeoff between the model accuracy and FL training delay. Concretely, a sophisticated model can fully utilize data resources for well-resourced IIoT devices, and thus the model accuracy is ameliorative. For limited-resourced IIoT devices, a relatively compact model can meet the requirement for low model accuracy but reduce the FL training delay.

Therefore, we present the ACT strategy to achieve the FL model match to different IIoT devices thus reducing the FL training delay while ensuring model accuracy. Concretely, the IIoT devices just train a proportion of the model parameters, and the proportion is determined by their local data volumes, and the adjusted parameters are determined by the Fisher information matrix. The workflow of the ACT strategy is as follows:
\subsubsection{Device Classification}
We perform clustering of the IIoT devices according to their local data volumes. Specifically, we use the following formulation as the distance evaluation functions\cite{arora2016analysis}:
\begin{equation}\label{eq:ACT1}
	\psi =\sum_{i=1}^{C} \sum_{n \in \mathcal{C}_{i}} \rVert D_{n}, m_{i}\rVert_2^2,
\end{equation}
where $C$ denotes the total number of clusters; $\mathcal{C}_i$ is the $i$-th cluster; and $m_i$ represents the medoid of the $i$-th cluster with $m_i \in \mathcal{C}_i$. The medoid $m_i$ can be computed by \cite{arora2016analysis}:
\begin{equation}\label{eq:ACT2}
	m_i=\underset{n \in \mathcal{C}_i}{\arg \min } \sum_{j \in \mathcal{C}_i} \rVert D_n, D_j \rVert_2^2.
\end{equation}

Hence, for each IIoT device $n$, its belonging cluster can be represented as a set $\mathcal{C}_i^n(1 \leq i \leq C)$.
\subsubsection{Parameters Proportion}
The proportion $\zeta_n$ of parameters that need to be trained for each IIoT device is also divided into $C$ classes according to their belonging clusters $\mathcal{C}_i^n$. For the same cluster of IIoT devices, the average of their data volumes is used to represent the data volume of the whole class, hence $\zeta_n$ can be calculated by:
\begin{equation}\label{eq:ACT4}
	\zeta_n = \frac{\operatorname{Mean}(\mathcal{C}_i^n)}{D_{\max}},
\end{equation}
where $\mathrm{Mean}(\mathcal{C}_i^n)$ represents the average value of all elements in $\mathcal{C}_i^n$; $D_\mathrm{max}$ is the max data volumes. Well-resourced IIoT devices are required to train more FL model parameters to maintain model accuracy, while limited-resourced devices focus on training fewer parameters of the FL model to reduce the FL training delay.

\subsubsection{Weights Freezing}
By employing the Fisher information matrix of the local FL model $\mathbf{w}_{n,t}$, we can estimate the importance of each parameter in the global FL model $\mathbf{w}_{g,t}$ relative to the local FL model $\mathbf{w}_{n,t}$. This can be expressed as:
\begin{equation}\label{eq:ACT6}
	\mathbf{I}_{g,t}^n = (\mathbf{w}_{g,t}-\mathbf{w}_{n,t})^\top \operatorname{diag}(\mathcal{I}_n(\mathbf{w}_{n,t}))(\mathbf{w}_{g,t}-\mathbf{w}_{n,t})
\end{equation}
where $\operatorname{diag}(\cdot)$ is the operation of obtaining diagonal elements; $\mathcal{I}_n(\mathbf{w}_{n,t})$ represents the Fisher information matrix of the local FL model $\mathbf{w}_{n,t}$. Then, a part of parameters of $\mathbf{I}_{g,t}^n$ with the smallest absolute value and their indexes are selected, and each IIoT device $n$ gets the corresponding global FL model $\mathbf{\hat{w}}_{g,t}^n$ according to the indexes $\mathbf{I}_{g,t,\mathrm{sel}}^n$. This can be expressed as follows:
\begin{equation}\label{eq:ACT8}
	\mathbf{I}_{g,t,\mathrm{sel}}^n = \operatorname{Select}(\mathbf{w}_{g,t},1-\zeta_n),
	\end{equation}

\begin{equation}\label{eq:ACT9}
	\mathbf{\hat{w}}_{g,t}^n = \operatorname{Freeze}(\mathbf{w}_{g,t},\mathbf{I}_{g,t,\mathrm{sel}}^n),
\end{equation}
where function $\operatorname{Select}(\mathbf{w}_{g,t},1-\zeta_n)$ represents the operation that select $1-\zeta_n$ proportion weights of the global FL model $\mathbf{w}_{g,t}$ with the smallest absolute values of $\mathbf{I}_{g,t}^n$ and output their indexes $\mathbf{I}_{g,t,\mathrm{sel}}^n$;
The function $\operatorname{Freeze}(\mathbf{w}_{g,t},\mathbf{I}_{g,t,\mathrm{sel}}^n)$ represents that in the global FL model $\mathbf{w}_{g,t}$, the weights of the selected parameters according to $\mathbf{I}_{g,t,\mathrm{sel}}^n$ are frozen.
The description of the ACT strategy is summarized in \textbf{Algorithm} \ref{alg:ACT}.

\subsubsection{Hessian Interpretation}
Since the Fisher information matrix approximates the Hessian $\mathbf{H}_{n,t}$ of the local loss $	{F_{n}}(\mathbf{w}_{n,t})$ at $\mathbf{w}_{n,t}=\mathbf{w}_{g,t}$ under some regularity conditions\cite{shoham2019overcoming}:
\begin{equation}\label{eq:ACT10}
	\mathbf{I}_{g,t}^n \approx\frac{1}{2} (\mathbf{w}_{g,t}-\mathbf{w}_{n,t})^\top \mathbf{H}_{n,t}(\mathbf{w}_{g,t}-\mathbf{w}_{n,t}),
\end{equation}
where $\mathbf{H}_{n,t}$ represents the Hessian matrix of $\mathbf{w}_{n,t}$, and Eq. (\ref{eq:ACT10}) is equal to the loss function. Therefore, the weights with small values in $\mathbf{I}_{g,t}^n$ can be frozen, and the weights with large values in $\mathbf{I}_{g,t}^n$ are exactly the goals we want to adjust to minimize the loss function.

\begin{algorithm}
	\caption{ACT strategy}
	\label{alg:ACT}
	\begin{algorithmic}[1]
		\REQUIRE $\mathbf{w}_{g,t}$.
		\ENSURE $\hat{\mathbf{w}}_{g,t}^n$.
		\STATE{Cluster the IIoT devices according to Eqs. (\ref{eq:ACT1}) and (\ref{eq:ACT2}), then each IIoT device $i$ obtains the belonging cluster $\mathcal{C}_i^n$.}
		\FOR{each IIoT device $n \in \mathcal{N}$ }
		\STATE{Calculate the trained parameters proportion $\zeta_n$  according to Eq. (\ref{eq:ACT4}).}
		\STATE{Obtain the specific global FL model $\mathbf{\hat{w}}_{g,t}^n$ for the IIoT device $n$ according to Eqs. (\ref{eq:ACT6}) - (\ref{eq:ACT9}).}
		\ENDFOR
	\end{algorithmic}
\end{algorithm}

In the ACT strategy, the partial weights of the global FL model are frozen to accommodate the local data volumes of the heterogeneous IIoT devices. This approach enables the local FL model to adapt better to the local data, thereby improving its accuracy. Simultaneously, the number of trainable local parameters in the IIoT devices is reduced, leading to a corresponding decrease in the FL training delay \cite{isikdogan2020semifreddonets,10269106}.

\subsection{Explainable SC Mechanism}
The inaccurate fire semantics may lead to incorrect decisions in fire relief. Hence, the explainability of extracted fire semantics is critical for being trustworthy \cite{liu2021trustworthy}. As a solution, the ESC mechanism is applied in IIoT devices to explain the extracted fire semantics from local FL models (namely the SC models). Specifically, we introduce a novel leakyReLU-based activation mapping to explain the relative between the semantic features extracted by the SC encoder and input monitoring data. The ESC mechanism employs semantic feature-specific gradient information to identify significant regions within input images. By visualizing this information, practitioners can gain insights into the underlying rationale for a particular prediction made by the SC model. As shown in Fig. \ref{fig:XSFL}(b), the steps of the ESC in a visual way are illustrated as follows:
\subsubsection{Semantic Gradient Computation}
We assume the semantic features $\mathbf{Q}=[Q_1,Q_2,..., Q_L]$ is a one-dimensional vector, where $Q_l$ represents weight value of the $l$-th feature.
Then, we compute the gradient of the semantic feature $Q_l$ to the feature map activation $A^k$ for the kernel $k$ of the last convolution layer, the calculated formulation is as follows:
\begin{equation}\label{eq:ESC1}
	G_{l}=\frac{\partial Q_l}{\partial A^k},
\end{equation}
where $G_{l}$ represents the gradients for the $l$-th semantic feature. In this step, note that all values calculated depend on the input monitoring data and the weights of the SC model are ﬁxed.

\subsubsection{Feature Significance Calculation}
We compute the neuron importance weights, $\omega_k^l$, by taking a weighted average of the pixel-wise gradients along both height $h$ and width $w$. The calculation of $\omega_k^l$ can be expressed as:
\begin{equation}\label{eq:ESC2}
	\omega_{k}^{l}=\sum_{i=1}^h \sum_{j=1}^w \varrho_{i,j}^{k,l} G_l,
\end{equation}
where $\varrho_{i,j}^{k,l}$ is the weighting coefficients for the pixel-wise gradients for semantic feature $Q_l$ and feature map $A^k$, which can be calculated by:
\begin{equation}\label{eq:ESC1-2}
	\varrho_{i,j}^{k,l}=\frac{\frac{\partial^{2} Q_{l}}{\left(\partial A_{i, j}^{k}\right)^{2}}}{2 \frac{\partial^{2} Q_{l}}{\left(\partial A_{i, j}^{k}\right)^{2}}+\sum\limits_{a=1}^h \sum\limits_{b=1}^w A_{a, b}^{k}\left\{\frac{\partial^{3} Q_l}{\left(\partial A_{i, j}^{k}\right)^{3}}\right\}}
\end{equation}
where $A_{i,j}^k$ and $A_{a,b}^k$ both represent someone pixel in $A^k$. Eq. (\ref{eq:ESC1-2}) shows that the weight of $A_{i,j}^k$ is affected by itself and the other pixels $A_{a,b}^k$.
Consequently, $\omega_{k}^{l}$ signifies a partial linearization of the deep network downstream from $A^k$, and encapsulates the importance of feature map $k$ concerning the $l$-th semantic feature \cite{selvaraju2016grad}.


\subsubsection{Visual Heatmap Generation}
The visual heatmap that ESC generated is a weighted combination of feature maps.
We introduce Leaky ReLU to retain the information of the negative values in the semantic features, thus solving the ``Dead Neuron" problem \cite{9219587} and ensuring the ESC mechanism generates remarkable heatmaps for the semantic explanation.
The localization map $P_l$ with Leaky ReLU for semantic feature $Q_l$ can be given by: 
\begin{equation}\label{eq:ESC3}
	\begin{split}
		&P_{l}=\operatorname{leakyReLU}\left(\sum_{e=1}^{k} \omega_{k}^{l} A^{k}\right)= \\
		&\max(0,\sum_{e=1}^{k} \omega_{k}^{l} A^{k}) + \alpha*\min(0,\sum_{e=1}^{k} \omega_{k}^{l} A^{k}),
	\end{split}
\end{equation}
where $\alpha\in(0,1)$ is the slope for negative
input. Finally, we obtain the complete localization map $\mathbf{P}$ for all semantics $\mathbf{S}$, where $\mathbf{P}=\{P_1,P_2,..., P_L\}$. Then, $\mathbf{P}$ is normalized and thus results in a heatmap for visualization. \textbf{Algorithm} \ref{alg:ESC} shows the description of the ESC mechanism.

\begin{algorithm}
	\caption{ESC mechanism}
	\label{alg:ESC}
	\begin{algorithmic}[1]
		\REQUIRE $\mathbf{S}$.
		\ENSURE $\mathbf{P}$.
		\FOR{each semantic feature $Q_l \in \mathbf{S}$}
		\STATE{Compute the gradient $G_l$ of the semantic feature $Q_l$} according to Eq. (\ref{eq:ESC1}). 
		\STATE{Obtain the neuron importance weights $\omega_k^l$ of the gradient $G_l$  according to Eqs. (\ref{eq:ESC2}) and (\ref{eq:ESC1-2}).}
		\STATE{Calculate $P_l$ according to Eq. (\ref{eq:ESC3}).}
		\ENDFOR
	\end{algorithmic}
\end{algorithm}

With the ESC mechanism, we achieve the explainability of the fire semantics from the SC model. The generated heatmap by ESC captures the relationship between the semantic features and input monitoring data, which helps the succors interpret the prediction results and thus improves the trustworthiness of SC.
In addition, the ESC mechanism can be used with the SC model locally on IIoT devices and without any pre-training, hence there is no extra computation cost for using ESC.

\section{Complexity Analysis}
Suppose the average data size of each IIoT device is denoted as $D_\text{avg}=\frac{1}{N}\sum_{n \in \mathcal{N}}D_n$. We assume the time complexity of communication and computation is linearly proportional to model sizes \cite{wu2022communication}. Then, the complexity analysis of the proposed XSFL in terms of communication and computation is performed.

In traditional FL, the communication cost primarily originates from uploading the local FL models during global aggregation. As a result, the communication time complexity is $O(T|\mathbf{w}_{g,t}|)$, where $|\cdot|$ denotes the operator for calculating the size of parameters. In the proposed XSFL, the communication time complexity is $O(T|\hat{\mathbf{w}}_{g,t}^n|)$. During the communication phase, since $|\hat{\mathbf{w}}_{g,t}^n|$ is equal to $|\mathbf{w}_{n,t}|$, our methods do not directly affect the communication complexity.

The time complexity of computation which depends on the local training in the traditional FL is $O(TGD_\text{avg}|\mathbf{w}_{g,t}|)$, where $T$ is the communication rounds, $G$ is the local training iterations of each IIoT device. In XSFL, partial parameters are frozen, namely, these parameters are not trained, hence the time complexity of computation is  $O(TGD_\text{avg}|\hat{\mathbf{w}}_{g,t}^{n,\mathrm{tr}}|)$, where $\hat{\mathbf{w}}_{g,t}^{n,\mathrm{tr}}$ means the trained weights of $\hat{\mathbf{w}}_{g,t}^{n}$. Due to $|\hat{\mathbf{w}}_{g,t}^{n,\mathrm{tr}}| < |\hat{\mathbf{w}}_{g,t}^{n}|$ and the weights freezing is carried on the ES, hence $O(TGD_\text{avg}|\hat{\mathbf{w}}_{g,t}^{n,\mathrm{tr}}|) < O(TGD_\text{avg}|\mathbf{w}_{g,t}|)$.

\section{Simulation and Discussion}

\subsection{Simulation Settings}
First, we use the Comprehensive Disaster Dataset (CDD) \cite{omar2020comparative}, which includes fire images, to evaluate the performance of XSFL and its competitors. The dataset is split into 80\% for training and 20\% for testing purposes. 

Second, we consider the following contenders:
\begin{itemize}
	\item FedAvg \cite{yang2019federated}: A benchmark FL algorithm.
	\item STC \cite{sattler2019robust}: A compression-based FL framework tailored to FL environment requirements.
	\item FTTQ \cite{xu2020ternary}: A ternary compression-based communication-efficient FL method.
	\item FedPAQ \cite{reisizadeh2020fedpaq}: A communication-efficient FL strategy incorporating periodic averaging and quantization.
	\item XSFL: Our proposed FL approach for IESN, as presented in this paper.
\end{itemize} 

Third, for the system model, we assume 10 IIoT devices are participating in FL training.
The CPU frequency for each IIoT device is set to $f_{n,t} = 2$ GHz.
Channel power gain is set at $h_{n,j}(t) = -50$ dB, and noise power is set to $\sigma = -174$ dBm/Hz.
Data transmission power and bandwidth are configured as $P^\text{U} = 0.01$ W, $P^\text{D} = 1$ W, $B^\text{U} = 1$ MHz, and $B^\text{D} = 20$ MHz \cite{chen2020joint}. 


\subsection{Evaluation of FL Model Performance}
In this subsection, we evaluate the proposed XSFL in terms of the model performance compared with other contenders. The evaluation indicators include Accuracy (ACC), Precision (PRE), Specificity (SPE), F1-score (F1), and Recall (REC) \cite{CHEN2023101803}. 
\begin{figure}[htbp]
	\centering
	\includegraphics[width=8.5cm]{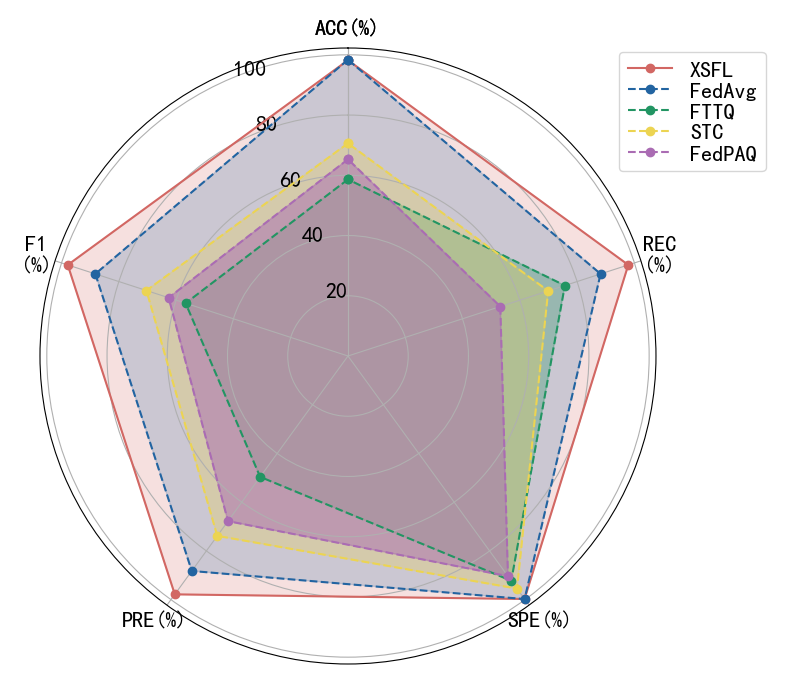}
	\caption{Model performance comparison among several FL schemes.}
	\label{fig:exp2}
\end{figure}

Based on the above settings, the final evaluation results of different FL schemes are shown in the radar plots of Fig. \ref{fig:exp2}. We can observe that the proposed XSFL achieves the best performance in all indicators.
We speculate the excellent performance of XSFL is attributed to the ACT strategy. ACT can freeze partial weights of the global FL model for adapting to the local data volumes of the IIoT devices, which assists the local FL model in making full use of the local data and thus improves the generalization of the global FL model.

\subsection{Evaluation of FL Training Delay}
This simulation aims to demonstrate the proposed XSFL can effectively reduce the FL training delay.
\begin{figure}[htbp]
	\centering
	\includegraphics[width=8.5cm]{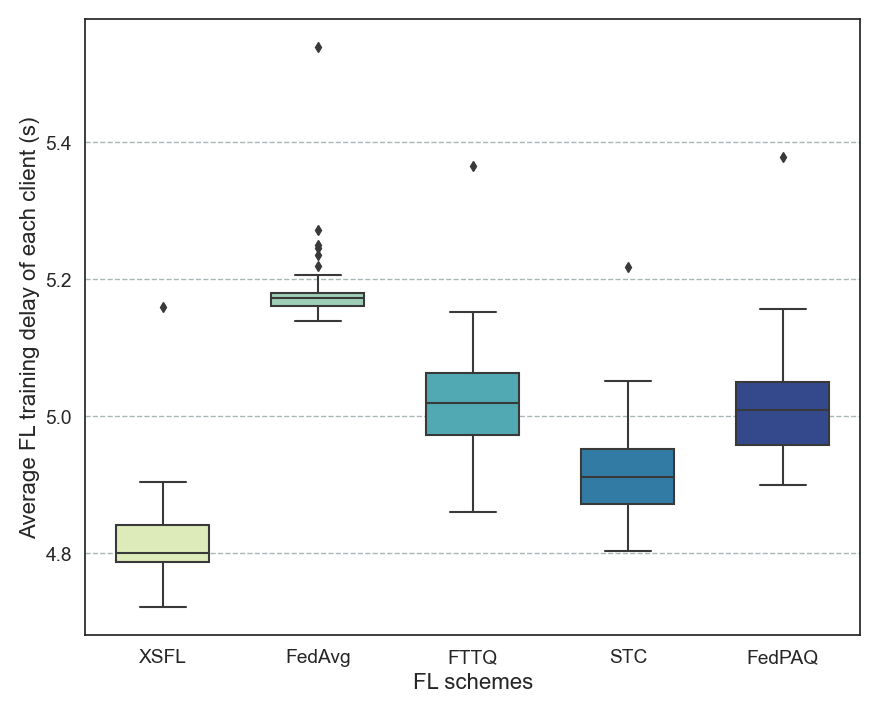}
	\caption{The average training delay of each IIoT device using several FL schemes.}
	\label{fig:exp4}
\end{figure}
\begin{figure}[htbp]
	\centering
	\includegraphics[width=8.5cm]{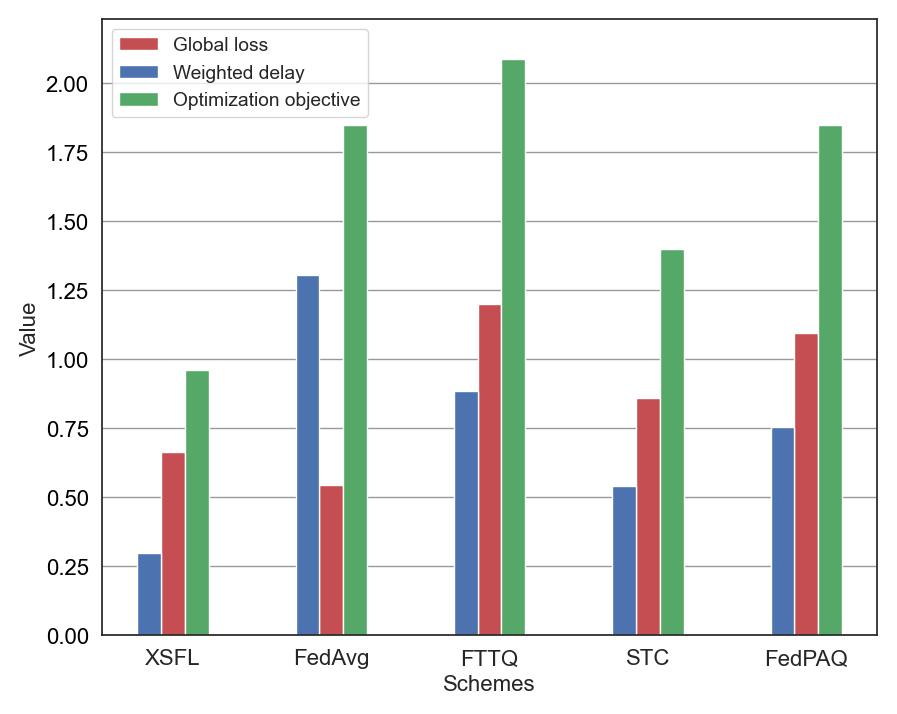}
	\caption{The values of the optimization objective for different FL schemes.}
	\label{fig:exp5}
\end{figure}

Fig. \ref{fig:exp4} shows the average FL training delay of each IIoT device in each communication round by several FL schemes. We can see that the boxplot of the presented XSFL scheme is at the bottom, which means the average training time of each IIoT device in XSFL is the shortest. 
Fig. \ref{fig:exp5} describes the global loss, weighted delay, and optimization objective under different FL schemes, where the weighted delay means the total training delay multiplied by a coefficient. The coefficient is set to 1e-3 to balance the delay and global loss; The optimization objective is the sum of the global loss and weighted delay. The results indicate that XSFL can effectively strike a balance between model performance and training delay in comparison to the other contenders.

\subsection{Evaluation of Semantic Explainability}
To evaluate the explainability of the XSFL, we display some ESC heatmaps corresponding to fire images from CDD. 
\begin{figure}[htbp]
	\centering
	\includegraphics[width=9cm]{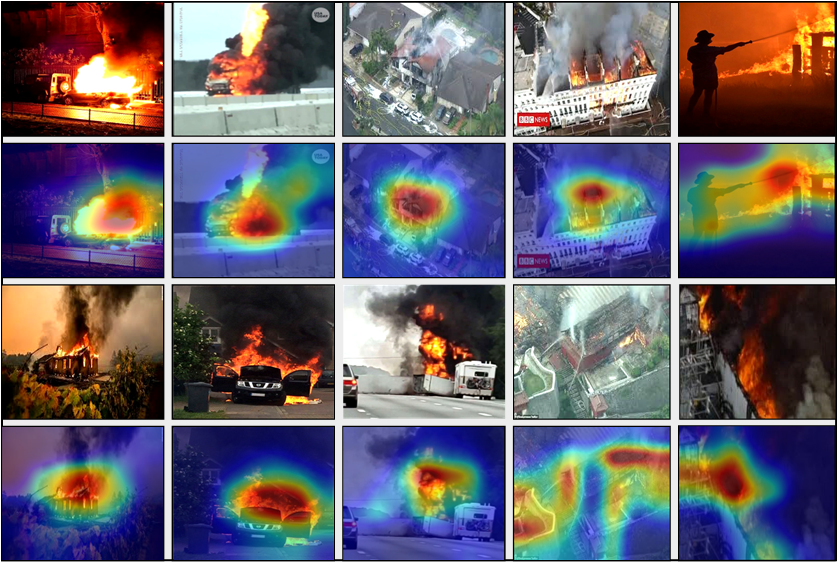}
	\caption{The explainable results of XSFL.}
	\label{fig:heatmap}
\end{figure} 

Fig. \ref{fig:heatmap} displays the heatmaps corresponding to the fire images, where the key features are accurately marked, such as the fire and smoke in the fire images. Through this visual explanation, the relationship between the semantic features and input monitoring images can be captured, which helps humans interpret the prediction results and improve the trustworthiness of SC. 

\section{Conclusions}
In this paper, IESN is presented to reduce the communication overheads of IIoT devices for fire surveillance. 
Furthermore, we propose the XSFL that enables IIoT devices and ES to collaboratively train the SC model while ensuring data privacy and security. In XSFL,
we first present the ACT strategy in ES, in which IIoT devices are clustered according to their local resources. Then, for each cluster, ES publishes a specific FL model that can better adapt to the device.
Second, the ESC mechanism is designed to provide explainability of the fire semantics, which helps humans understand the semantic results and improve the trustworthiness of SC.
Finally, the simulation results demonstrate the superiority of the proposed XSFL method. 
 
\bibliographystyle{ieeetran}
\bibliography{bare_jrnl_bobo}

\begin{thebibliography}{10}
\providecommand{\url}[1]{#1}
\csname url@samestyle\endcsname
\providecommand{\newblock}{\relax}
\providecommand{\bibinfo}[2]{#2}
\providecommand{\BIBentrySTDinterwordspacing}{\spaceskip=0pt\relax}
\providecommand{\BIBentryALTinterwordstretchfactor}{4}
\providecommand{\BIBentryALTinterwordspacing}{\spaceskip=\fontdimen2\font plus
\BIBentryALTinterwordstretchfactor\fontdimen3\font minus
  \fontdimen4\font\relax}
\providecommand{\BIBforeignlanguage}[2]{{%
\expandafter\ifx\csname l@#1\endcsname\relax
\typeout{** WARNING: IEEEtran.bst: No hyphenation pattern has been}%
\typeout{** loaded for the language `#1'. Using the pattern for}%
\typeout{** the default language instead.}%
\else
\language=\csname l@#1\endcsname
\fi
#2}}
\providecommand{\BIBdecl}{\relax}
\BIBdecl

\bibitem{prabha2019iot}
B.~Prabha, ``An iot based efficient fire supervision monitoring and alerting
  system,'' in \emph{2019 Third International conference on I-SMAC (IoT in
  Social, Mobile, Analytics and Cloud)(I-SMAC)}.\hskip 1em plus 0.5em minus
  0.4em\relax IEEE, 2019, pp. 414--419.

\bibitem{thekkil2017real}
T.~M. Thekkil and N.~Prabakaran, ``Real-time wsn based early flood detection
  and control monitoring system,'' in \emph{2017 International Conference on
  Intelligent Computing, Instrumentation and Control Technologies
  (ICICICT)}.\hskip 1em plus 0.5em minus 0.4em\relax IEEE, 2017, pp.
  1709--1713.

\bibitem{jiang2020ai}
F.~Jiang, K.~Wang, L.~Dong, C.~Pan, W.~Xu, and K.~Yang, ``{AI} driven
  heterogeneous mec system with uav assistance for dynamic environment:
  Challenges and solutions,'' \emph{IEEE Network}, vol.~35, no.~1, pp.
  400--408, 2020.

\bibitem{muhammad2019efficient}
K.~Muhammad, S.~Khan, M.~Elhoseny, S.~H. Ahmed, and S.~W. Baik, ``Efficient
  fire detection for uncertain surveillance environment,'' \emph{IEEE
  Transactions on Industrial Informatics}, vol.~15, no.~5, pp. 3113--3122,
  2019.

\bibitem{9531399}
A.~Jain and A.~Srivastava, ``Privacy-preserving efficient fire detection system
  for indoor surveillance,'' \emph{IEEE Transactions on Industrial
  Informatics}, vol.~18, no.~5, pp. 3043--3054, 2022.

\bibitem{manikandan2022design}
G.~Manikandan, D.~Karunkuzhali, D.~Geetha, and V.~Kavitha, ``Design of an iot
  approach for security surveillance system for industrial process monitoring
  using raspberry-pi,'' in \emph{AIP Conference Proceedings}, vol. 2519,
  no.~1.\hskip 1em plus 0.5em minus 0.4em\relax AIP Publishing LLC, 2022, p.
  030024.

\bibitem{10084949}
S.~Banka, B.~Kanchanapalli, B.~Annam, A.~Akula, S.~Thirukkovalluru, and
  S.~Jarpula, ``A novel method for fire detection and alarming notification
  using iot technology,'' in \emph{2023 Second International Conference on
  Electronics and Renewable Systems (ICEARS)}, 2023, pp. 560--565.

\bibitem{10558819}
F.~Jiang, Y.~Peng, L.~Dong, K.~Wang, K.~Yang, C.~Pan, and X.~You, ``Large ai
  model-based semantic communications,'' \emph{IEEE Wireless Communications},
  vol.~31, no.~3, pp. 68--75, 2024.

\bibitem{yang2022privacy}
F.~Yang, Y.~Qiao, M.~Z. Abedin, and C.~Huang, ``Privacy-preserved credit data
  sharing integrating blockchain and federated learning for industrial 4.0,''
  \emph{IEEE Transactions on Industrial Informatics}, vol.~18, no.~12, pp.
  8755--8764, 2022.

\bibitem{chen2013natural}
D.~Chen, Z.~Liu, L.~Wang, M.~Dou, J.~Chen, and H.~Li, ``Natural disaster
  monitoring with wireless sensor networks: A case study of data-intensive
  applications upon low-cost scalable systems,'' \emph{Mobile Networks and
  Applications}, vol.~18, no.~5, pp. 651--663, 2013.

\bibitem{yang2019federated}
Q.~Yang, Y.~Liu, Y.~Cheng, Y.~Kang, T.~Chen, and H.~Yu, ``Federated learning,''
  \emph{Synthesis Lectures on Artificial Intelligence and Machine Learning},
  vol.~13, no.~3, pp. 1--207, 2019.

\bibitem{begoli2019need}
E.~Begoli, T.~Bhattacharya, and D.~Kusnezov, ``The need for uncertainty
  quantification in machine-assisted medical decision making,'' \emph{Nature
  Machine Intelligence}, vol.~1, no.~1, pp. 20--23, 2019.

\bibitem{wang2022transformer}
Y.~Wang, Z.~Gao, D.~Zheng, S.~Chen, D.~Gunduz, and H.~V. Poor,
  ``Transformer-empowered {6G} intelligent networks: From massive mimo
  processing to semantic communication,'' \emph{IEEE Wireless Communications},
  2022.

\bibitem{park2020end}
S.~Park, O.~Simeone, and J.~Kang, ``End-to-end fast training of communication
  links without a channel model via online meta-learning,'' in \emph{2020 IEEE
  21st International Workshop on Signal Processing Advances in Wireless
  Communications (SPAWC)}.\hskip 1em plus 0.5em minus 0.4em\relax IEEE, 2020,
  pp. 1--5.

\bibitem{10398266}
L.~Dong, F.~Jiang, M.~Wang, Y.~Peng, and X.~Li, ``Deep progressive
  reinforcement learning-based flexible resource scheduling framework for irs
  and uav-assisted mec system,'' \emph{IEEE Transactions on Neural Networks and
  Learning Systems}, pp. 1--13, 2024.

\bibitem{gunduz2022beyond}
D.~G{\"u}nd{\"u}z, Z.~Qin, I.~E. Aguerri, H.~S. Dhillon, Z.~Yang, A.~Yener,
  K.~K. Wong, and C.-B. Chae, ``Beyond transmitting bits: Context, semantics,
  and task-oriented communications,'' \emph{IEEE Journal on Selected Areas in
  Communications}, vol.~41, no.~1, pp. 5--41, 2022.

\bibitem{wei2021low}
K.~Wei, J.~Li, C.~Ma, M.~Ding, C.~Chen, S.~Jin, Z.~Han, and H.~V. Poor,
  ``Low-latency federated learning over wireless channels with differential
  privacy,'' \emph{IEEE Journal on Selected Areas in Communications}, vol.~40,
  no.~1, pp. 290--307, 2021.

\bibitem{10091843}
X.~Zhou, X.~Ye, K.~I.-K. Wang, W.~Liang, N.~K.~C. Nair, S.~Shimizu, Z.~Yan, and
  Q.~Jin, ``Hierarchical federated learning with social context
  clustering-based participant selection for internet of medical things
  applications,'' \emph{IEEE Transactions on Computational Social Systems},
  vol.~10, no.~4, pp. 1742--1751, 2023.

\bibitem{arora2016analysis}
P.~Arora, S.~Varshney \emph{et~al.}, ``Analysis of k-means and k-medoids
  algorithm for big data,'' \emph{Procedia Computer Science}, vol.~78, pp.
  507--512, 2016.

\bibitem{shoham2019overcoming}
N.~Shoham, T.~Avidor, A.~Keren, N.~Israel, D.~Benditkis, L.~Mor-Yosef, and
  I.~Zeitak, ``Overcoming forgetting in federated learning on non-iid data,''
  \emph{arXiv preprint arXiv:1910.07796}, 2019.

\bibitem{isikdogan2020semifreddonets}
L.~F. Isikdogan, B.~V. Nayak, C.-T. Wu, J.~P. Moreira, S.~Rao, and G.~Michael,
  ``Semifreddonets: Partially frozen neural networks for efficient computer
  vision systems,'' in \emph{European Conference on Computer Vision}.\hskip 1em
  plus 0.5em minus 0.4em\relax Springer, 2020, pp. 193--208.

\bibitem{10269106}
X.~Zhou, X.~Zheng, X.~Cui, J.~Shi, W.~Liang, Z.~Yan, L.~T. Yang, S.~Shimizu,
  and K.~I.-K. Wang, ``Digital twin enhanced federated reinforcement learning
  with lightweight knowledge distillation in mobile networks,'' \emph{IEEE
  Journal on Selected Areas in Communications}, vol.~41, no.~10, pp.
  3191--3211, 2023.

\bibitem{liu2021trustworthy}
H.~Liu, Y.~Wang, W.~Fan, X.~Liu, Y.~Li, S.~Jain, Y.~Liu, A.~K. Jain, and
  J.~Tang, ``Trustworthy ai: A computational perspective,'' \emph{arXiv
  preprint arXiv:2107.06641}, 2021.

\bibitem{selvaraju2016grad}
R.~R. Selvaraju, A.~Das, R.~Vedantam, M.~Cogswell, D.~Parikh, and D.~Batra,
  ``Grad-cam: Why did you say that?'' \emph{arXiv preprint arXiv:1611.07450},
  2016.

\bibitem{9219587}
J.~Xu, Z.~Li, B.~Du, M.~Zhang, and J.~Liu, ``Reluplex made more practical:
  Leaky relu,'' in \emph{2020 IEEE Symposium on Computers and Communications
  (ISCC)}, 2020, pp. 1--7.

\bibitem{wu2022communication}
C.~Wu, F.~Wu, L.~Lyu, Y.~Huang, and X.~Xie, ``Communication-efficient federated
  learning via knowledge distillation,'' \emph{Nature communications}, vol.~13,
  no.~1, pp. 1--8, 2022.

\bibitem{omar2020comparative}
A.~Omar, S.~Ashraf, A.~Rahman, M.~A. Amin, A.~A. Ali \emph{et~al.}, ``A
  comparative study on disaster detection from social media images using deep
  learning,'' in \emph{Proceedings of the Global AI Congress 2019}.\hskip 1em
  plus 0.5em minus 0.4em\relax Springer, 2020, pp. 485--499.

\bibitem{sattler2019robust}
F.~Sattler, S.~Wiedemann, K.-R. M{\"u}ller, and W.~Samek, ``Robust and
  communication-efficient federated learning from non-iid data,'' \emph{IEEE
  transactions on neural networks and learning systems}, vol.~31, no.~9, pp.
  3400--3413, 2019.

\bibitem{xu2020ternary}
J.~Xu, W.~Du, Y.~Jin, W.~He, and R.~Cheng, ``Ternary compression for
  communication-efficient federated learning,'' \emph{IEEE Transactions on
  Neural Networks and Learning Systems}, 2020.

\bibitem{reisizadeh2020fedpaq}
A.~Reisizadeh, A.~Mokhtari, H.~Hassani, A.~Jadbabaie, and R.~Pedarsani,
  ``Fedpaq: A communication-efficient federated learning method with periodic
  averaging and quantization,'' in \emph{International Conference on Artificial
  Intelligence and Statistics}.\hskip 1em plus 0.5em minus 0.4em\relax PMLR,
  2020, pp. 2021--2031.

\bibitem{chen2020joint}
M.~Chen, Z.~Yang, W.~Saad, C.~Yin, H.~V. Poor, and S.~Cui, ``A joint learning
  and communications framework for federated learning over wireless networks,''
  \emph{IEEE Transactions on Wireless Communications}, vol.~20, no.~1, pp.
  269--283, 2020.

\bibitem{CHEN2023101803}
Z.-S. Chen, J.-Y. Chen, Y.-H. Chen, Y.~Yang, L.~Jin, E.~Herrera-Viedma, and
  W.~Pedrycz, ``Large-group failure mode and effects analysis for risk
  management of angle grinders in the construction industry,''
  \emph{Information Fusion}, vol.~97, p. 101803, 2023.

\end{thebibliography}
\section*{Biographies}

\vspace{-10mm}
\begin{IEEEbiography}[{\includegraphics[width=1in,height=1.25in,keepaspectratio]{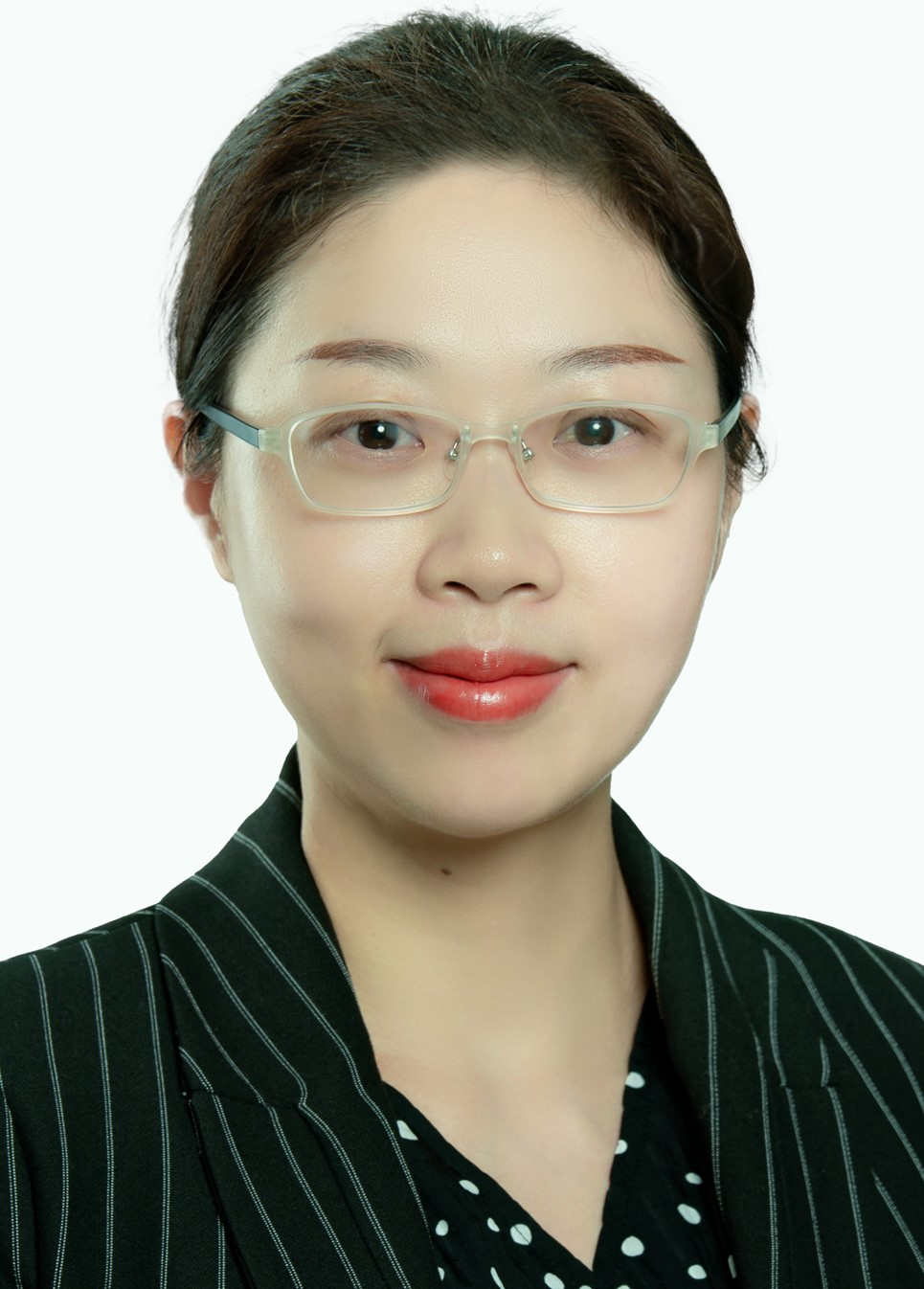}}]{Li Dong} received the B.S. and M.S. degrees in School of Physics and Electronics from Hunan Normal University, China, in 2004 and 2007, respectively. She received her Ph.D. degree in School of Geosciences and Info-physics from the Central South University, China, in 2018. She is currently an Associate Professor with the Hunan University of Technology and Business, Changsha. Her research interests include federated learning, Internet of Things, and mobile edge computing.
\end{IEEEbiography}

\vspace{-10mm}
\begin{IEEEbiography}[{\includegraphics[width=1in,height=1.25in,keepaspectratio]{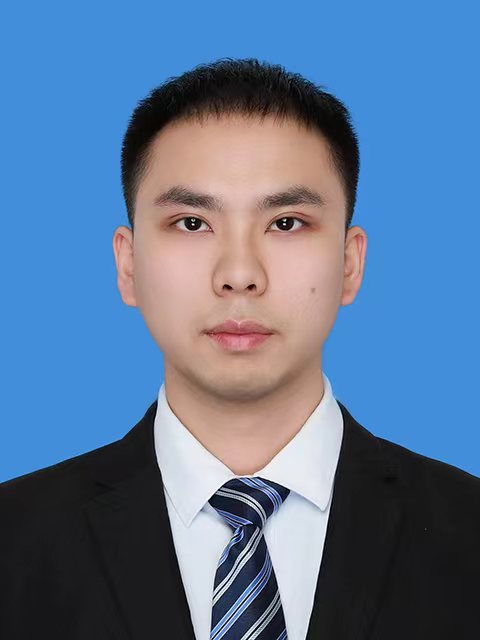}}]{Yubo Peng} received his B.S. and M.S. degrees from Hunan Normal University, Changsha, China, in 2019 and 2024. He is pursuing a doctor’s degree from the School of Intelligent Software and Engineering at Nanjing University. His main research interests include semantic communication and large models.
\end{IEEEbiography}
\vspace{-10mm}
\begin{IEEEbiography}[{\includegraphics[width=1in,height=1.25in,keepaspectratio]{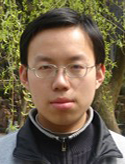}}]{Feibo Jiang} received his B.S. and M.S. degrees in School of Physics and Electronics from Hunan Normal University, China, in 2004 and 2007, respectively. He received his Ph.D. degree in School of Geosciences and Info-physics from the Central South University, China, in 2014. He is currently an associate professor at the Hunan Provincial Key Laboratory of Intelligent Computing and Language Information Processing, Hunan Normal University, China. His research interests include semantic communication, federated learning, Internet of Things, and mobile edge computing.
\end{IEEEbiography}

\vspace{-10mm}
\begin{IEEEbiography}[{\includegraphics[width=1in,height=1.25in,keepaspectratio]{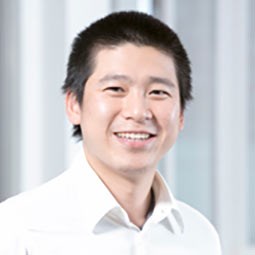}}]{Kezhi Wang}received the Ph.D. degree in Engineering from the University of Warwick, U.K. He was with the University of Essex and Northumbria University, U.K. Currently, he is a Senior Lecturer with the Department of Computer Science, Brunel University London, U.K. His research interests include wireless communications, mobile edge computing, and machine learning.
\end{IEEEbiography}

\vspace{-10mm}
\begin{IEEEbiography} [{\includegraphics[width=1in,height=1.25in,clip,keepaspectratio]{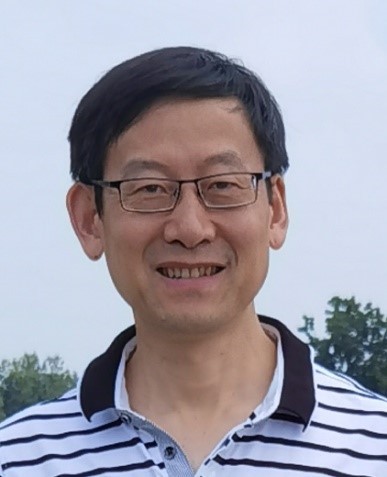}}]{Kun Yang} received his PhD from the Department of Electronic \& Electrical Engineering of University College London (UCL), UK. He is currently a Chair Professor in the School of Intelligent Software and Engineering, Nanjing University, China. He is also an affiliated professor of University of Essex, UK. His main research interests include wireless networks and communications, communication-computing cooperation, and new AI (artificial intelligence) for wireless. He has published 400+ papers and filed 30 patents. He serves on the editorial boards of a number of IEEE journals (e.g., IEEE WCM, TVT, TNB). He is a Deputy Editor-in-Chief of IET Smart Cities Journal. He has been a Judge of GSMA GLOMO Award at World Mobile Congress – Barcelona since 2019. He was a Distinguished Lecturer of IEEE ComSoc (2020-2021). He is a Member of Academia Europaea (MAE), a Fellow of IEEE, a Fellow of IET and a Distinguished Member of ACM.
\end{IEEEbiography} 
\newpage
\end{document}